\documentclass{article}
\usepackage{ijcai16}

\usepackage{times}

\usepackage{amsfonts}
\usepackage{mathtools,amsmath}
\usepackage{graphicx}

\title{Trading-Off Cost of Deployment Versus Accuracy in Learning Predictive Models
  \thanks{This project was supported by NSF IIS-1418590 and the Johns Hopkins University IDIES Seed Funding Initiative.}}
\author{Daniel P. Robinson \thanks{Equal contribution.} \\ 
Johns Hopkins University \\
Department of Applied Mathematics and Statistics \\
Baltimore, Maryland  \\
daniel.p.robinson@jhu.edu \\
\And
Suchi Saria \footnotemark[2] \\
Johns Hopkins University \\
Department of Computer Science \\
Baltimore, Maryland \\
ssaria@cs.jhu.edu
}

\begin{document}

\maketitle

\begin{abstract}
Predictive models are finding an increasing number of applications in many industries.
As a result, a practical means for trading-off the cost of deploying a model versus 
its effectiveness is needed. Our work is motivated by risk prediction
problems in healthcare. Cost-structures in domains such as healthcare are quite complex, posing a significant challenge to existing approaches. We propose a novel framework for designing cost-sensitive structured regularizers that is suitable for problems with
complex cost dependencies. We draw upon a surprising connection to boolean circuits. In particular,
we represent the problem costs as a multi-layer boolean circuit, and
then use properties of boolean circuits to define an extended feature
vector and a group regularizer that exactly captures the underlying cost
structure. The resulting regularizer may then be
combined with a fidelity function to perform model prediction, for example.  
For the challenging real-world application of risk prediction for sepsis in intensive
care units, the use of our regularizer leads to models that are in harmony with the underlying cost structure and thus provide an excellent prediction accuracy versus cost tradeoff.
\end{abstract}

\makeatletter
\newcommand{\band}{{\rm AND}}
\newcommand{\bor}{{\rm OR}}
\newcommand{\bnot}{{\rm NOT}}
\newcommand{\bgets}{\hookleftarrow}
\newcommand{\words}[1]{\mgap\text{#1}\mgap}
\newcommand{\Ascr}{{\mathcal A}}
\newcommand{\Bscr}{{\mathcal B}}
\newcommand{\Fscr}{{\mathcal F}}
\newcommand{\Mscr}{{\mathcal M}}
\newcommand{\Nscr}{{\mathcal N}}
\newcommand{\Sscr}{{\mathcal S}}
\newcommand{\sub}[1]{^{\null}_{#1}}
\newcommand{\lamclock}{\lambda_{time}}
\newcommand{\mytinyclock}{time}
\newcommand{\strt}{\rule[-.5ex]{0pt}{3ex}}
\providecommand{\diag} {\mathop{\operator@font{diag}}}
\newcommand{\into}{\rightarrow}
\newcommand{\sgap}{\;}
\newcommand{\mgap}{\;\;}
\newcommand{\bgap}{\;\;\;}
\newcommand{\T}{^T\!}
\newcommand{\drop}{^{\null}}
\newcommand{\norm}[1]{\|#1\|}
\newcommand{\onenorm}[1]{\norm{#1}_1\drop}
\newcommand{\minim}{\mathop{\operator@font{minimize}}}
\newcommand{\minimize}[1]{{\displaystyle\minim_{#1}}}
\newtheorem{remark}{Remark}
\makeatother

\section{Introduction}
\begin{figure*}
\centering
\includegraphics[scale=0.45]{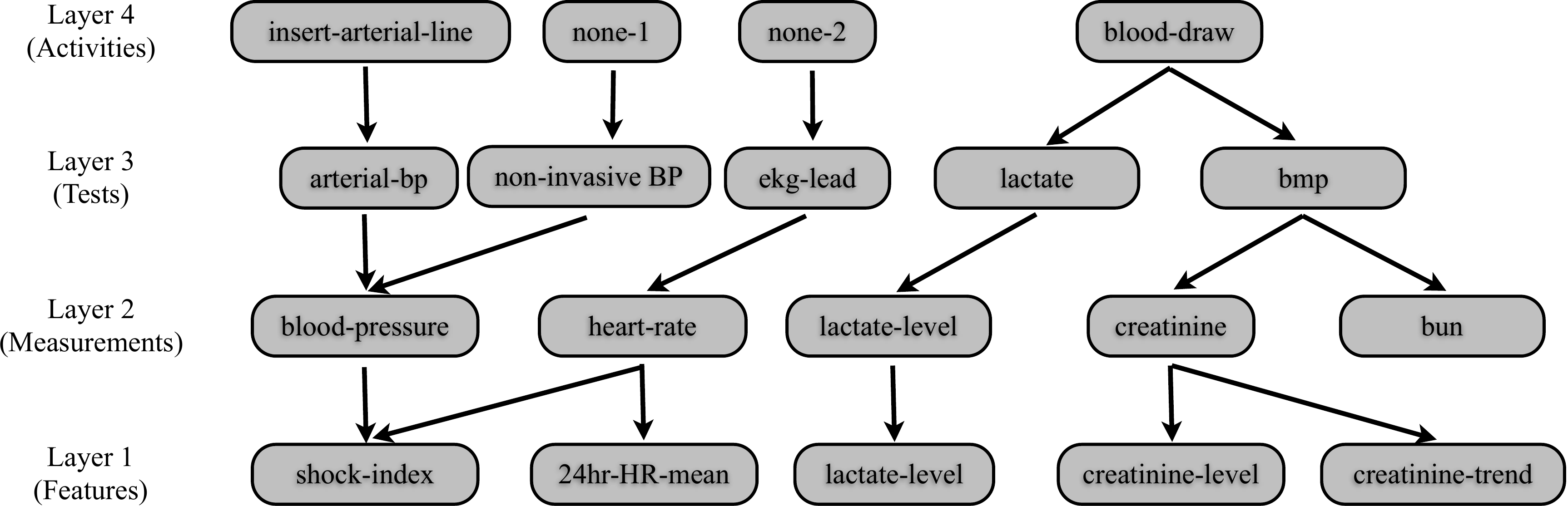}
\caption{A portion of the dependency graph for the ICU example.}
\label{fig:graph}
\end{figure*}

Many industries (e.g., retail, manufacturing, and
medicine) are recognizing the advantages of using predictive models
to make key decisions. They also understand that the cost of obtaining input measurements
should be balanced with their effectiveness in prediction when choosing which model to deploy. This is especially challenging when the cost structure for an application is complicated. As an important example, consider the cost structure associated with deploying a predictive model in an Intensive Care Unit (ICU) (see the cost-dependency graph in Figure~\ref{fig:graph}).
In such a setting, the following hold: (i) costs may be defined for tests, measurements, or activities and these costs may be of different types (e.g., the financial cost of acquiring a blood test versus the staff time taken to draw blood); (ii) features are obtained using one or more measurements (e.g., lactate level or creatinine) which in turn are obtained by ordering a test; (iii) a test may consist of a single measurement
(e.g., lactate level) or a panel of measurements (e.g., CBC panel);
(iv) a measurement can be ordered via
multiple tests (e.g., creatinine can be ordered on its own, as
part of a basic or a comprehensive metabolic panel, each having a
different financial cost); (v) multiple features can be derived from the same
measurement (e.g., the heart rate variability and the heart rate trend can both be derived from the heart rate trace); and (vi) some features 
may require multiple measurements (e.g., shock index
is derived from blood pressure and heart rate measurements).
These aspects make the cost structure complicated.

Before expanding upon the challenges involved with addressing complex cost-structures such as the one above, we first introduce the mathematical setup for learning predictive models. 
This involves data that is
formally represented by sets of pairs $\{(x_i,y_i)\}_{i=1}^N$ for
some integer $N$, where $x_i\in\Re^n$ for some
integer $n$ and $y_i\in\Re$, for all $1 \leq i \leq N$. The vector
$x_i$ denotes the 
$i$th input (feature) vector and $y_i$ the output (label)
associated with the $i$th input vector $x_i$.  
The goal is then to predict the \emph{unknown} output associated with
a newly obtained input vector by
using the knowledge one learns from the data
$\{(x_i,y_i)\}_{i=1}^N$.  A popular approach for performing this
task is to build predictive models via empirical regularized-loss
minimization~\cite{vapnik1998statistical}.  The problems used take the form 
\begin{equation} \label{prob:opt}
\minimize{\beta\in\Re^n} \mgap
f(\beta) := \frac{1}{N}\sum_{i=1}^N L\big(\beta; (x_i,y_i)\big) + R(\beta),
\end{equation}
where $\beta\in\Re^n$ is the parameter vector to be learned, $L$ is a loss function such as  
the logistic loss $L(\beta;(x_i,y_i)) = \log(1+\exp^{y_ix_i\T \beta})$, 
and $R$ is a regularizer.
The choice of regularizer amounts to
giving preference to certain models, e.g., the
$\ell_1$-regularizer $R(\beta) = \lambda\onenorm{\beta}$ for some $\lambda > 0$ prefers models defined by a sparse 
vector $\beta$. In practice, 
the regularizer should be chosen to reflect
the preferred models, which are often driven by the costs associated with the application.
For example, in compressed sensing, one wishes to find a sparse solution to a linear
system of equations.  Thus, the cost, i.e., the number of nonzeros in
a prospective solution, is
harmonious with the $\ell_1$-regularizer, which promotes sparse
solutions.  Note that in this example, as well as many others, 
the costs are \emph{directly} tied to the feature vectors themselves, i.e., they occur at the feature level. 

How does one design an appropriate regularizer for problems with a complicated cost structure, such as for the ICU example above? We address that question in this paper. \\

\noindent\textbf{Related work.}
Learning models in the presence of costs has received significant attention in recent years (e.g., \cite{ICML2012Xu_592,Ji:2007:CFA:1224549.1224576,weiss2013learning,xu2013cost,raykar2010designing}). Existing work has primarily targeted applications where the cost of \emph{computation} is the primary concern, and this cost is elicited at the feature level. Moreover, much of this work has focused on optimizing performance when information is acquired incrementally \cite{Ji:2007:CFA:1224549.1224576,ICML2012Xu_592,trapeznikov2013supervised,kanani2008prediction,kapoor2009breaking}. In~\cite{Ji:2007:CFA:1224549.1224576}, they define the problem of cost-sensitive classification and use a partially observable Markov decision process to trade-off the cost of acquiring additional measurements with classification performance. While they apply their method to a medical diagnosis problem, their costs were approximated at the feature level. 
In~\cite{ICML2012Xu_592}, stage-wise regression is used to learn a collection of regression trees in a manner that ensures that classifiers built from more trees is more accurate, but more expensive. For the task of ranking web page documents, they showed improved speed and accuracy by accounting for feature costs---simple lookups (e.g., word occurrences) versus those needing more computation (e.g., a document-specific BM25 score for measuring relevance to a query). For structured prediction, \cite{weiss2013dynamic} proposed a two tier 
collection of models of varying costs and a model selector; for each new test example, their selector adaptively chooses a model. For vision applications (e.g., articulated pose estimation in videos), they showed gains in performance by adaptively selecting models of varying costs by using a histogram of gradient features at a fine (expensive) versus a coarse (cheap) resolution.  These solutions focused on applications with no dependencies between the costs for the units reasoned over (i.e. feature or model costs are independent) and when they are provided upfront. As predictive models continue to find their way into many important applications, 
a means for incorporating rich cost structures is needed.
  
Returning to our example in healthcare, the challenge of incorporating costs arises from the dependencies between features, measurements, tests, and required activities.  Measurements may be obtained from a singleton test or as part of a test that yields multiple measurements. Tests may have different resource costs associated with them, while features may be derived from more than one measurement. These dependencies between features, measurements and tests yield a complex dependence structure between the features. Moreover, various costs are specified at different levels of this hierarchy; therefore, the cost of a feature is not specified upfront, but rather is dependent on which other features, measurements, and tests are selected. 

Cost imposed via a hierarchical dependency graph is reminiscent of past works utilizing structured sparsity penalties (see the survey~\cite{wainwright2014structured} and \cite{bach2012optimization}), especially those using tree-based regularizers \cite{kim2010tree} and penalties with overlapping groups and hierarchical structure \cite{zhao2006grouped,bach2012optimization}. Different from these past works, a key challenge for our task is that the structure of the group regularizer is not given and its construction is not straightforward. We show that cost-dependency graphs are naturally captured via Boolean circuits---graphs where nodes share a combination of AND and OR connections with its parents. Only leaf nodes (i.e. feature nodes) of this circuit are included in the regularizer while the internal nodes (e.g., measurements needed to obtain features) induce dependency between the leaf nodes. The presence of mixed AND/OR relationships and the non-inclusion of internal nodes renders our application different from past works. Other regularizers such as OSCAR \cite{zhong2012efficient} and Laplacian Net~\cite{huang2011sparse} aim at discovering group structure when the features are highly correlated. In our setting, the groups are determined by the structure of the cost graph, not by the correlations between the features. \\

\noindent\textbf{Our contributions.}
We develop a new framework for defining
structured regularizers suitable for problems with complex
cost structures by drawing upon a surprising connection to boolean circuits.
In particular, we represent the problem costs as a boolean circuit, and then use properties of boolean circuits to define the \emph{exact cost penalty}.
 Based on our exact cost penalty, any standard convex relaxation may be employed for the purpose of computational efficiency, and here we choose a standard $\ell_1$-$\ell_\infty$ relaxation.
Our new regularizer may be used within
an empirical risk minimization framework to tradeoff cost versus
accuracy. We focus on the one-shot setting (i.e., when all measurements are obtained upfront), although
our regularizer is also applicable in the incremental setting.
Since the cost-structure of many real-life applications may be represented as a boolean circuit, the contribution of our work is substantial.

Our ideas are presented in the context of a
challenging healthcare application---the development of a rapid screening
tool for sepsis \cite{SepsisNEJM}---using data from patients in the
ICU~\cite{lit:MIMICII_ref_2}. In this setting, examples of potential users include patients, doctors, and administration. Our experiments show
that our regularizer
allows for a collection of
models that are in harmony with a user's cost preferences.
Numerical comparisons to a cost-sensitive $\ell_1$---a natural competitor to our proposed regularizer that does not account for the complicated cost structure---shows that models obtained with our regularizer have a better prediction/cost tradeoff. Compared to existing approaches in predictive modeling where cost preferences are often accounted for post hoc, our scheme provides a new way to account for complex cost preferences during model selection.

\section{Regularizers for complex cost structures}

Our scheme is general since it may be applied to any problem with a cost
structure that may be represented as a finite-layer boolean circuit. 
However, for clarity of exposition, 
we first focus on a particular healthcare application that also
serves as the basis for the numerical results presented. \\

\subsection{An example from the intensive-care unit (ICU)}
We formulate a structured regularizer for the cost structure associated with
risk prediction applications for the in-hospital setting.  These
include problems such as prediction of those at
risk for death, the likelihood of readmission, and the early detection
of adverse events, e.g., shock and cardiac arrest. 

Recall the cost dependency graph for the ICU example in Figure~\ref{fig:graph}.
The features are represented by nodes in
layer-$1$, and their calculation requires a subset of measurements 
from layer $2$, i.e.,
nodes in layer-$1$ share an \band{} or \bor{} relationship with those
  in layer-$2$.
Measurements can be
obtained in a number of ways by performing various tests, which are
represented at layer-$3$, i.e.,
nodes in layer-$2$ similarly share an \band{} or \bor{} relationship with those
  in layer-$3$.
The caregiver activities are
represented at layer-$4$ and are performed when a test is
needed that requires that action, i.e., layer-$3$ shares an \band{}
relationship with layer-$4$. Every relationship in this
boolean circuit is described using only logical
\band{} and \bor{} operations. Note that, without loss of
generality, we include
fictitious nodes ``none-1'' and ``none-2'' in layer-$4$ so that the collection
of input nodes are in the same layer.

There are three relevant costs:
the financial cost of ordering a test, the waiting time 
to obtain a test result, and the caregivers' time needed to
perform the activities required for the tests. The ideal regularizer should account
for the following:
(i) obtaining a measurement may cost different amounts
  depending on which tests are ordered to obtain it;
(ii) features share costs with other features derived from the same
  measurement;
(iii) a feature may require multiple measurements so that its
  cost depends on more than one measurement; and
(iv) caregiver time and financial costs are additive while
  wait time is the maximum of the separate wait times.

Our structured regularizer requires the following sets:
\begin{alignat*}{2}
\mathcal{F} &:= \{f_1, \cdots, f_{n_f}\},  \mgap &
\bgap \mathcal{M} &:= \{m_1, \cdots, m_{n_m}\}, \mgap \\
\mathcal{T} &:= \{t_1, \cdots, t_{n_t}\}, \mgap & 
\mathcal{A} &:= \{a_1, \cdots, a_{n_a}\}.
\end{alignat*}
to be the sets of features (layer-$1$ nodes), measurements (layer-$2$
nodes), tests (layer-$3$ nodes), and caregiver activities (layer-$4$
nodes), with $n_f,n_m, n_t,$ and $n_a$ being the number of each, respectively.
We use $f_i \bgets m_i$ to mean that there is a directed
edge that links node $m_j$ to node $f_i$. Thus, 
our specific boolean circuit allow
us to interpret  
$f_i \bgets m_j$,
$m_j \bgets t_k$, and $t_k \bgets a_l$
to mean the $i$th feature requires the $j$th measurement, the $j$th measurement can be obtained by
performing the $k$th test, and the $k$th test requires the $l$th activity.

We now define the set valued mappings
$m(f_i) := \{m_j: f_i \bgets m_j  \}$,
$t(m_j) := \{t_k: m_j \bgets t_k  \}$, and
$t(a_l) := \{t_k: t_k \bgets a_l  \}$,
which represent the set of measurements
required to obtain feature $i$, the set of tests that
produce measurement $j$, and the set of tests that require
action $l$. We note that we have overloaded the
definition of the function $t$ above, i.e., we have two different definitions for
$t(m_j)$ and $t(a_l)$. However, this should not lead to any confusion
since the correct definition is always clear from the context.  

Next, we resolve the fact that some features may be
obtained in multiple ways by ordering various combinations of tests.
If this is not considered, the cost of a feature may  
be over penalized by our regularizer.  
To address this issue, let $w_i$ denote the numbers of ways feature
$i$ can be obtained.
Then, for the $i$th feature, we define
$
\vec{f}_i := [f_{i,1},\cdots,f_{i,w_i}]^T$
and
$\vec{\beta}_i := [\beta_{i,1},\cdots,\beta_{i,w_i}]^T
$
so that $\beta_{i,p}$ represents the parameter associated with
ordering feature $f_i$ in the $p$th way. 
This allows us to define
the \emph{extended} feature and parameter vectors 
$\vec{f} := [\vec{f_1}, \dots, \vec{f_{n_f}}]^T$
and $\vec{\beta} := [\vec{\beta_1}, \dots, \vec{\beta_{n_f}}]^T$. \\

\noindent\textbf{Modeling financial cost and caregiver time:} 
To model financial cost, which is incurred at
the test level, for each test $t_k$ and feature $f_i$ we define
$
\vec{n}_{k,i} := [n_{k,i,1}, \cdots, n_{k,i,w_i}]^T
$
with
$$
n_{k,i,p} :=\! 
\begin{cases} 
1 & \!\mbox{if $t_k$ is used when $f_i$ is ordered in its $p$th way,} \\
0 & \!\mbox{otherwise,}
\end{cases}
$$
for all
$1 \leq k \leq n_t$, 
$1 \leq i \leq n_f$, and
$1 \leq p \leq w_i$.
Given a financial cost $C_k^{\mathcal{T}}$ of ordering test $k$ 
and a weighting parameter $\lambda_{\$}$,
the \emph{exact} structured regularizer for financial cost is 
\begin{equation}  \label{P-exact}
   R_{exact}^{\$}(\vec{\beta})
   := \lambda_{\$} \!\sum_{k=1}^{n_t} \mathcal{C}_k^{\mathcal{T}} I\!\left( \sum_{i=1}^{n_f} \sum_{p=1}^{w_i}
   I(n_{k,i,p}\beta_{i,p}) \right)
\end{equation} 
with the indicator function $I$ satisfying $I(0) = 0$ and $I(z) = 1$ for
all $z \neq 0$.
It follows from~\eqref{P-exact} that a financial cost
for test $t_k$ is incurred only when
instructed to order some feature $f_i$ in the $p$th
way ($\beta_{i,p} \neq 0$), and that $p$th way requires test $t_k$
($n_{k,i,p} \neq 0$). 
The regularizer~\eqref{P-exact} is not computationally friendly, so we instead use the relaxed structured regularizer 
\begin{equation}
\begin{aligned}
  R_{relax}^{\$}(\vec{\beta})
  &:= \lambda_{\$} \!\sum_{k=1}^{n_t} \mathcal{C}_k^{\mathcal{T}} 
    \Big\| 
    \bigvee_{i=1}^{n_f} \vec{n}_{k,i} \odot \vec{\beta}_i  
    \Big\|_{\infty}
    \label{P-approx}
\end{aligned}
\end{equation}
where we define for a set of vectors $\{z_i\}_{i=1}^{n}$ and
a subset $S = \{i_1,i_2,\dots, i_r\} \subseteq \{1,2,\dots,n\}$ the quantities
$$
\bigvee_{i=1}^{n} z_i := 
[ z_1^T \dots z_{n}^T ]^T
\words{and}
\bigvee_{i\in\Sscr} z_i := 
[ z_{i_1}^T \dots z_{i_r}^T ]^T.
$$
Note that~\eqref{P-approx} is a sum of group
$\ell_\infty$-norms, which is supported by the software SPAMS.

To model the caregivers time cost in a similar way, we define 
$
\vec{n}_{l,i} := [n_{l,i,1}, \cdots, n_{l,i,w_i}]^T
$
with
$$
n_{l,i,p} :=\! 
\begin{cases} 
1 & \!\mbox{if $a_l$ is used when $f_i$ is ordered in its $p$th way,} \\
0 & \!\mbox{otherwise,}
\end{cases}
$$
for all
$1 \leq l \leq n_a$, 
$1 \leq i \leq n_f$, and
$1 \leq p \leq w_i$,
where we have again overloaded notation. The regularizer associated
with caregiver activity time then becomes
\begin{equation}
\begin{aligned}
  R_{relax}^{\mytinyclock}(\vec{\beta})
  &:= \lamclock \!\sum_{l=1}^{n_a} \mathcal{C}_l^{\mathcal{A}} 
    \Big\| 
    \bigvee_{i=1}^{n_f} \vec{n}_{l,i} \odot \vec{\beta}_i  
    \Big\|_{\infty} 
    \label{P-approx-activity}
\end{aligned}
\end{equation}
with $\mathcal{C}_l^{\mathcal{A}}$ being the time cost associated with the $l$th activity and
$\lamclock > 0$ a weighting parameter. Overall, our
structured regularizer becomes
\begin{equation} \label{def:R-relaxed}
R_{relax}(\vec{\beta}) : = R_{relax}^{\$}(\vec{\beta}) + R_{relax}^{\mytinyclock}(\vec{\beta}).
\end{equation}
By varying $\lambda_{\$}$ and $\lamclock$ we trade-off the
financial and caregiver activity time costs, respectively.
\begin{remark}
If a scaled-$\ell_1$-norm was used, the user chooses a weight for each
feature by condensing the complex cost structure into a single
number, necessarily in an ad-hoc way.
\end{remark}
\begin{remark}\label{remark:circuit}
Consider the $3$-layer boolean circuit where 
layer-$1$ contain the nodes $\Fscr$,
layer-$2$ contain the nodes $\mathcal{Z} := \{ f_{i,p} : 1 \leq i \leq
n_f \ \text{and} \ 1 \leq p \leq w_i\}$, and
layer-$3$ contain the nodes $\Ascr$. Moreover, the gate functions at
layer-$1$ are given, for each $f_i$, by
$
g_{f_i}(\mathcal{Z}) 
:= \underset{1 \leq p \leq w_i}{\bor} f_{i,p}
$
for all $1 \leq i \leq n_f$, and the gate functions at layer-$2$ are
given, for each $f_{i,p}$, by
$
g_{f_{i,p}}(\Ascr)
:= \underset{\{l: n_{l,i,p} = 1\} }{\band} a_l
$ 
for all $1 \leq i \leq n_f$ and $1 \leq p \leq w_i$.  In particular,
only \bor{} gate functions are used in layer-$1$ and only \band{} gate
functions are used in layer-$2$.  Moreover, the properties of this
$3$-layer gate allows us to conclude that for a given
caregiver activity, say $a_l$, we have
$
\left\| 
\bigvee_{i=1}^{n_f} \vec{n}_{l,i} \odot \vec{\beta}_i  
\right\|_{\infty}
\equiv \left\| \bigvee_{(i,p)\in\Sscr_l} \vec{\beta}_{i,p} \right\|_{\infty}
$
with the index set $\Sscr_l$ defind as
$\Sscr_l 
:= \{ (i,p) : n_{l,i,p} = 1\} 
= \{ (i,p) : \, \text{the output of $g_{f_{i,p}}(\cdot)$ depends on $a_l$} \}$,
so that the definition of our regularizer~\eqref{P-approx-activity}
follows from our knowledge of the $3$-layer boolean circuit. In fact,
the only properties of the circuit that we used were (i) layer-$1$ was
the feature layer; (ii) layer-$3$ contained the nodes whose 
costs we were modeling; (iii) layer-$1$ only contained \bor{}
gates; and (iv) layer-$2$ only had \band{} gates. This motivates the
general case below.
\end{remark}

\noindent\textbf{Modeling testing wait time:} 
We use a simpler approach to address the time needed to obtain test results.
Note that the wait time for a set of test results is 
the maximum of the wait times for each individual test
(assuming that tests can be ordered in parallel). Thus, for a
given upper bound, say $W$, on the tolerated testing wait time, we
only allow tests that have a wait time less than $W$
to be used. This amounts to selecting a reduced boolean circuit
containing only these allowed tests, the caregiver actions required to obtain these
allowed tests, measurements that result from the allowed tests, and
the features that may be calculated from the included measurements. \\

\subsection{Structured regularizer: the general case}
We now show how to define our regularizer for any problem whose cost structure may be represented as a finite $r$-layer boolean circuit; Figure~\ref{fig:graph} is an instance of such a circuit.

An $r$-layer boolean circuit 
consists of layers of finitely many nodes. The lowest layer (layer-$1$) consists of the set of output nodes, while the highest layer (layer-$r$) contains the input nodes. Additionally, we are given
boolean functions---defined on the basis $\Bscr = \{ \band,\bor,\bnot \}$---for all nodes.  Formally, each boolean function performs the basic logical operations from $\Bscr$ on one or more logical inputs from the previous layer, and produces a single logical output value.  The healthcare example in Figure~\ref{fig:graph} is a $4$-layer boolean circuit with the features corresponding to layer-$1$, the measurements to layer-$2$, the tests to layer-$3$, and the activities to layer-$4$.

Let $\Nscr_i := \{x_{i,1},x_{i,2}, \dots, x_{i,n_i}\}$
be the nodes in layer-$i$ for some $n_i$. By removing double negations, and using the laws of distribution and De Morgan's laws, the $r$-layer circuit may be reduced to a $3$-layer boolean circuit in disjunctive normal form~\cite{pfahringer2010conjunctive,zeng2012reduction}.  The nodes in the $3$-layer circuit are then
layer-$3$: $\{x_{r,1},x_{r,2}, \dots, x_{r,n_r}\}$,
layer-$2$: $\{z_{1},z_{2}, \dots, z_{m}\}$, and
layer-$1$: $\{x_{1,1},x_{1,2}, \dots, x_{1,n_1}\}$
for some $m$ and set $\{z_i\}_{i=1}^m$ of nodes for layer-$2$.
Moreover, the only logical operations used by the boolean functions $g_{z_i}(\cdot)$ in layer-2 are \band{} and \bnot{} operations, while the boolean functions $g_{x_{1,i}}(\cdot)$ in layer-$1$ only use logical \bor{} operations.  (In Remark~\ref{remark:circuit} we showed how a circuit of this form could be obtained for the healthcare example.)
If we define the vectors
$\vec{z} := [z_{1}, z_{2}, \dots, z_{m}]^T$
and
$\vec{\beta} :=  [\beta_{1}, \beta_{2}, \dots, \beta_{m}]^T$,
then we define our cost-driven structured regularizer as
\begin{equation*}    
  R\sub{relax}(\,\vec{\beta}\,)
  := \lambda \!\sum_{k=1}^{n_r} \mathcal{C}_k 
    \Big\| 
    \bigvee_{j\in\Sscr_{k}} \beta_j
    \Big\|_{\infty} 
\end{equation*}
with $\Sscr_{k} := 
   \{ j : g_{z_j}(\cdot) \, \text{depends on the logical value of} \,  x_{r,k}\}$.
When this regularizer is used in model prediction, an optimal value for the extended vector  $\vec{\beta}$ is obtained.  Using this vector and the fact that layer-$1$ only has \bor{} gates, we know that a node $x_{1,i}$ in layer-$1$ (i.e., the feature layer for the healthcare application) has the logical value of $1$ (i.e., the feature should be computed) if  $\beta_j \neq 0$ for some $j \in \{ k : g_{x_{1,i}}(\cdot) \ \text{depends on the logical value of} \ z_{k}\}$.

\begin{remark} Although our exact penalty is approximated by an overlapping group regularizer, what is non-trivial is determining which features belong to which groups for complex cost graphs. By relating the cost graph to a Boolean circuit, we can use properties of Boolean circuits to define an extended feature set and overlapping structure that is correct for arbitrary cost graphs. Moreover, this connection allows for the use of widely used off-the-shelf software such as SymPy
to convert an arbitrary graph to the 3-layer circuit in disjunctive normal form used to define our exact regularizer.
\end{remark}


\section{Numerical experiments}\label{sec:numerical}
We focus on early detection of septic shock---an adverse event
resulting from sepsis---since it is the 11th leading cause of patient mortality in the United
States. (Mortality rates are between $30\%$ and $50\%$ for those who develop
septic shock~\cite{lit:angus_infection}.) Although early treatment 
can reduce the patient mortality rate, less than one-third of
patients receive appropriate therapy before onset. Therefore,
an early warning system that accurately predicts a sepsis event 
allows for appropriate treatment and a higher quality of patient care.
(See the references in \cite{Ho2014:TMIS} for recent work
on sepsis detection; none have tackled the cost of deployment.)
More broadly, this problem is an instance of cost-sensitive risk prediction for automated triage~\cite{wilson1981computerized}. 

\begin{figure*}[t]
\begin{center}
\includegraphics[scale=0.63]{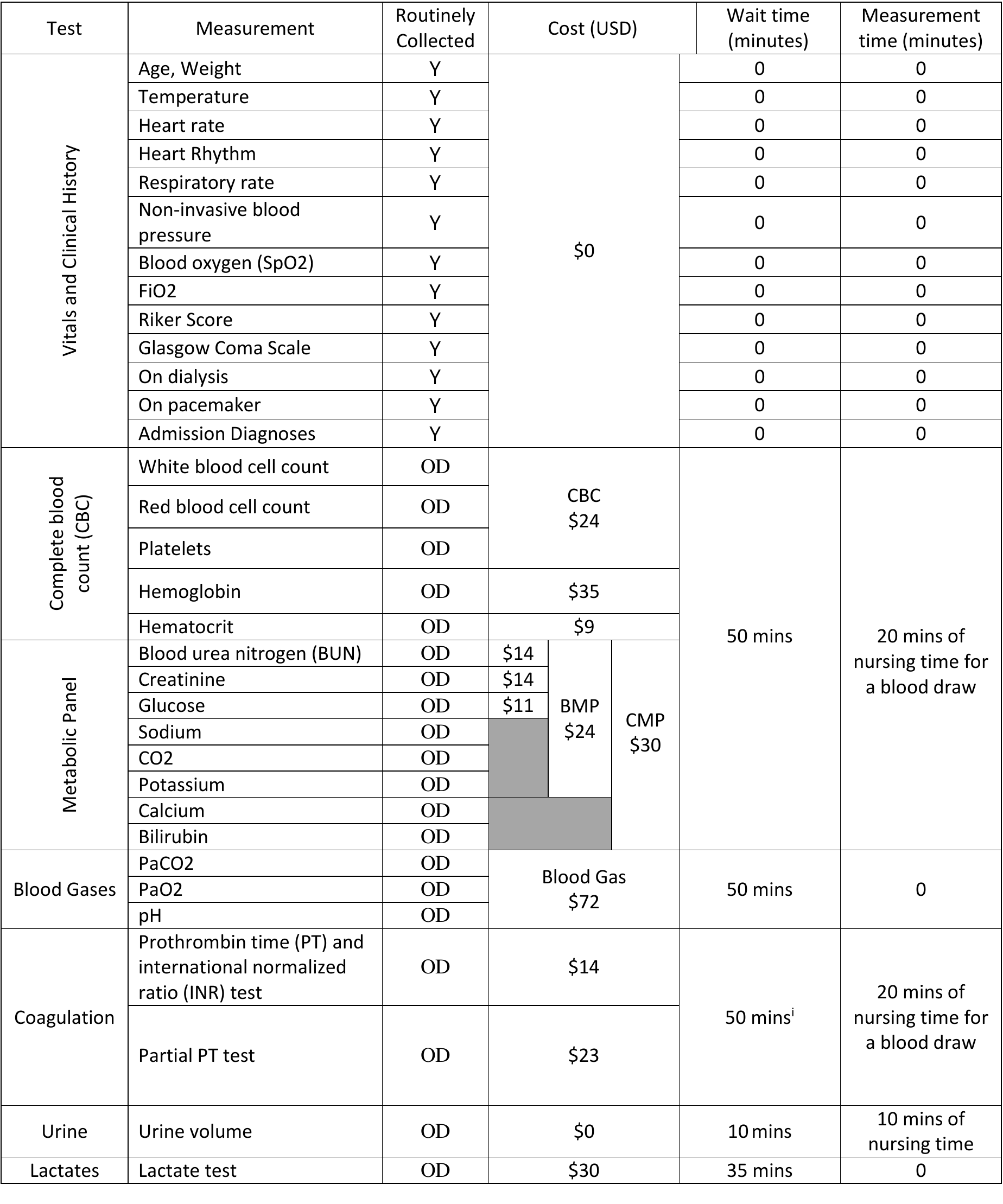}
 \end{center}
\caption{The cost structure (measurements and associated costs) for risk
prediction of adverse events in an ICU.  
Some measurements are made on demand (OD), while the others are routinely collected.  The following acronyms are used: basic metabolic panel (BMP), comprehensive metabolic panel (CMP), and complete blood count (CBC).} 
\label{fig:measurements}
\end{figure*}

\begin{table*}[t]
\caption{Various costs for different models obtained by using our structured regularizer.}\label{tab:aucs}
\small
\begin{center}
  \begin{tabular}{|c||c|c|c|c|}\hline
      \strt Models              & $\mathcal{M}_1$ & $\mathcal{M}_2$ & $\mathcal{M}_3$ & $\mathcal{M}_4$                                   \\ \hline
    \strt Sensitivity at $0.85$ specificity        & 0.61 & 0.66 & 0.65 & 0.72
                              \\ \hline
    \strt AUC                 & $82.79\pm 0.55$ & $84.45\pm0.64$ & $84.75\pm0.55$  & $87.21\pm0.46$                                      \\ \hline
    \strt Financial Cost      & \$$0$           & \$$0$          & \$$72$          & \$$170$                                             \\ \hline
    \strt Caregiver Time      & 0 minutes       & $10$ minutes   & 0 minutes       & $30$ minutes                                        \\ \hline
    \strt Result Time         & 0 minutes       & $10$ minutes   & 50 minutes      & $50$ minutes                                        \\ \hline
    \strt Tests Required      & routine         & routine, urine          & abg, routine            & abg, cbc, cmp, hct, hemoglobin, routine, urine \\ \hline
    \strt Activities Required & none            & urine          & arterial stick & arterial stick, blood draw, urine                   \\ \hline
  \end{tabular}
\end{center}
\end{table*}

We constructed the full cost-graph in collaboration with domain experts, which resulted in $119$ nodes and $294$ edges. (The full list of measurements and tests can be found in Figure~\ref{fig:measurements}.)
We combine the logistic-regression function with our
structured regularizer~\eqref{def:R-relaxed} to predict the
probability that a patient will develop septic shock. We use
MIMIC-II~\cite{lit:MIMICII_ref_2}, a large publicly available dataset
of electronic health records from patients admitted to four
different ICUs at the Beth Israel Deaconess Medical Center over a seven year
period. Using the processing described in \cite{henry2015targeted},
$2,\!291$ positive patients and $12,\!646$ negative patient cases were
obtained.

We answer two questions.  Does our structured regularizer lead to diverse models, especially in terms of the various costs? How well does our new structured regularizer perform compared to existing available solutions? Natural comparisons include regularizers that account for cost but do not account for the cost-dependence structure, e.g., the $\ell_1$-norm regularizer or the cost-sensitive $\ell_1$-norm. 
A comparison with other structured sparsity penalties would also seem appropriate, but none exist that construct the penalty for complex cost graphs (see the discussion in the related work section).  We do not include comparisons to stage-wise alternatives because they are suboptimal to the proposed cost-sensitive $\ell_1$-norm, which yields a global optimum.   \\

\noindent\textbf{Experimental setup.} 
We split the individuals 
into training (75\%) and test (25\%) sets. From the training set,
we process the data using a sliding window to extract positive and
negative samples consisting of the features observed at a given time,
and an associated label that is positive if septic shock was
occured within the following $48$ hours and negative
otherwise. Since the dataset is imbalanced, we subsample the
negative pairs to obtain a balanced training set.

For our test set, we use the learned model to predict
the risk of septic shock at each time point. 
This gives a trajectory of risk for septic shock over time for
each individual. For a given threshold, an individual is 
\emph{identified} as having shock if their risk
trajectory rose above that threshold prior to shock onset. For
this threshold, we calculate: (i) sensitivity as the fraction of
patients who develop septic shock and are identified as having a high
risk of septic shock; (ii) the false positive rate (FPR) as the
fraction of patients who never develop septic shock but are identified
as high risk patients by our model; and specificity as $1 -
\text{FPR}$. The receiver operating characteristic (ROC) curve and 
area under that curve (AUC)
are obtained by varying the threshold value, with patients identified
as at-risk if their predicted probability was above the threshold
value. We use $10$ bootstrapped samples to estimate confidence intervals for the AUC. 

We used the {\sc mexFistGraph} routine in SPAMS to minimize the sum of
the logistic function and our structured regularizer~\eqref{def:R-relaxed}.
The maximum allowed iteration limit was set to $5,\!000$ and the termination tolerance (duality gap) to $10^{-3}$. \\

\noindent\textbf{Model diversity.}
Three costs were considered: (i) financial cost associated with ordering a test; (ii) nursing-staff's time needed to perform the activities required for the tests; and (iii) waiting time to obtain a test result. For a chosen maximum wait time and weighting parameters $\lambda_{\$}$ and $\lamclock$,
our algorithm minimizes the sum of the logistic-regression function with the regularizer~\eqref{def:R-relaxed}, which returns parameters for a model from which we may compute an associated ROC, AUC, financial cost, nurse-time, and test result wait time. By sweeping over a range of values for the maximum allowed wait time, $\lambda_{\$}$, and $\lamclock$,
we obtain models with various costs that reflect preferences for different models.  For our cost-dependency structure,
there are three possible maximum wait times: $50$ minutes, $10$ minutes, and $0$ minutes. For each of these scenarios, we select values for $\lambda_{\$}$ and $\lamclock$
from an equally spaced grid over the interval $[10^{-3}, 10^{-7}]$, which yields a collection of models at the cost-accuracy frontier. Four models---denoted as $\Mscr_1$, $\Mscr_2$, $\Mscr_3$, and $\Mscr_4$---are represented in Table \ref{tab:aucs} to illustrate the tradeoff achieved by our approach.

Model $\mathcal{M}_1$ is the most cost-effective.
It  uses existing measurements that are routinely collected and
therefore it neither incurs a financial cost nor the need for
nursing-time to acquire new measurements. Since no additional tests
are required, the wait time for the model is also zero minutes. 
The model achieves a relatively high AUC of $82.79$.
The set of measurements that were
found to be most predictive include: clinical history (on ventilator,
on pacemaker, has cardiovascular complications); vitals (shock index,
raw and derived features of the heart rate, SpO2, FiO2, blood
pressure, respiratory rate); and time since first presentation of
systemic inflammatory response syndrome (SIRS). 

At the other extreme, model $\mathcal{M}_4$ has a financial cost of $\$170$, requires a nurse-time of $30$ minutes, and a total test result wait time of $50$ minutes. It requires measurements attained from numerous additional tests such as the arterial blood gas, comprehensive metabolic panel, hematocrit, hemoglobin, and urine tests. By using these measurements, the accuracy increases to an AUC of $87.21$, and  
shows a clinically significant gain in sensitivity compared to model $\mathcal{M}_1$.

Models $\mathcal{M}_2$ and $\mathcal{M}_3$ have cost and performance intermediate to models $\Mscr_1$ and $\Mscr_4$. Also, it is interesting to see that $\mathcal{M}_2$ and $\mathcal{M}_3$ achieve similar performance in very different ways.  Model $\Mscr_2$ selects a urine measurement with a test result wait time of $10$ minutes and $10$ minutes of nurse time, while $\Mscr_3$ does not require any nurse time, but needs $50$ minutes of wait time to receive test results.

For the specificity level of $0.85$, the models vary significantly in terms of sensitivity.  As expected, model $\Mscr_1$ has the lowest sensitivity value of $0.61$, followed by model $\Mscr_3$ with a value of $0.65$, then model $\Mscr_2$ with a value of $0.66$, and finally model $\Mscr_4$ with a value of $0.72$.  Thus, with additional resources, $\Mscr_4$ is significantly better at
identifying patients that eventually did experience septic shock. The
added sensitivity is useful for units with vulnerable populations. 

In practice, a user can benefit from our structured regularizer in at least two ways.  First, the user can obtain multiple predictive models by choosing a diverse set of values for the weighting parameter values $\lambda_{\$}$ and $\lamclock$.
This brute force approach would provide a diverse landscape of models with very different cost distributions.  A second approach involves the user making a sequence of decisions.  In particular, the user would adjust the weighting parameter values after the results using their current values is obtained. Specifically, the user would adjust the parameter values so as to obtain a new model that is more aligned with their preferences. \\

\begin{figure}
   \includegraphics[width=0.95\linewidth]{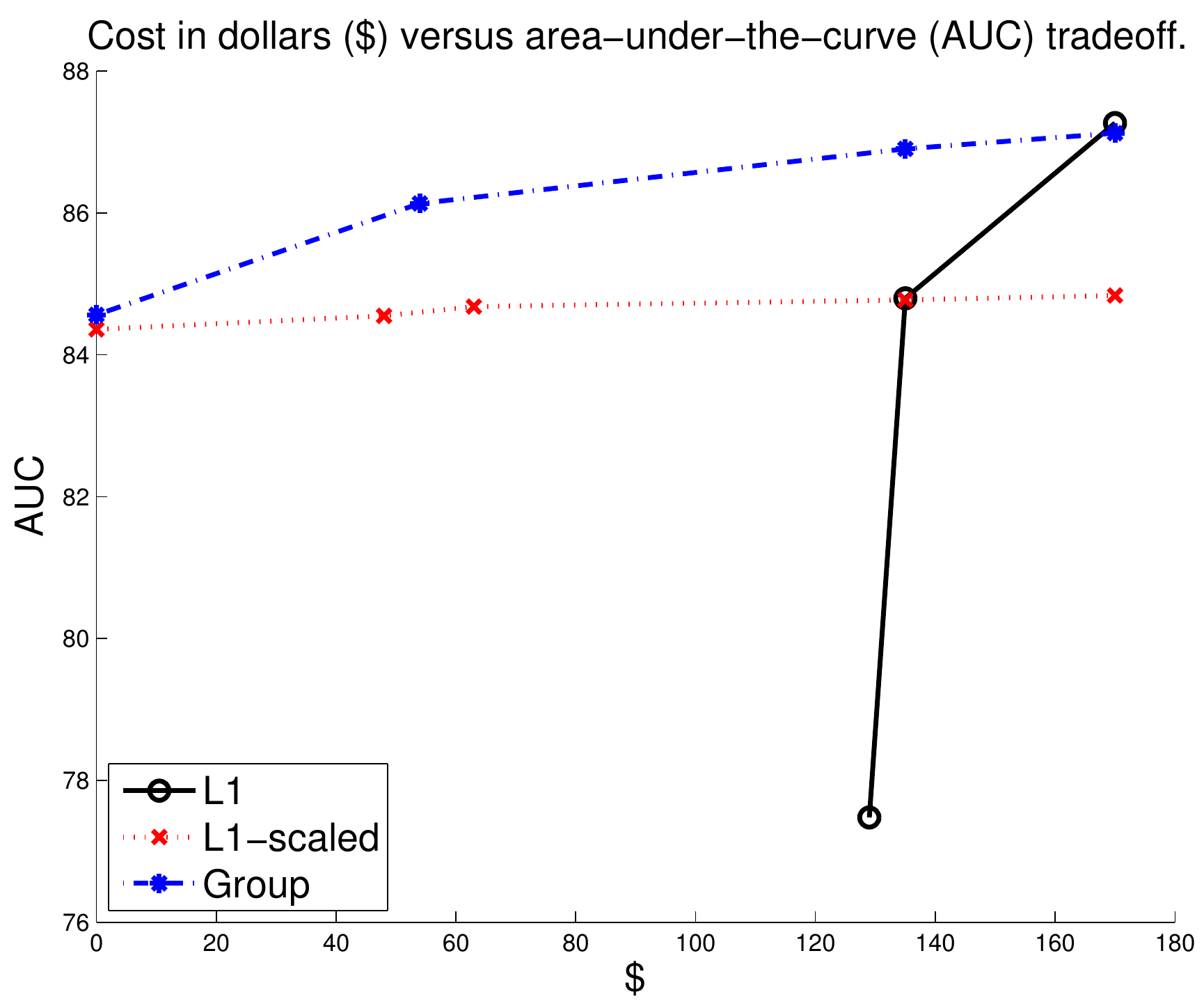}
   \caption{Cost (\$) and area under the curve (AUC) for the $\ell_1$ (L1), scaled-$\ell_1$ (L1-scaled), and group regularizer (Group).}
   \label{fig:l1-comparison}
\end{figure}

\noindent\textbf{Comparison with the $\ell_1$ and scaled $\ell_1$-norm.}
Simple regularizers (e.g., the $\ell_1$-norm) can not capture the rich structure of the cost-dependencies in real-world 
domains such as healthcare.
Figure~\ref{fig:l1-comparison} compares our structured group regularizer (Group) to the $\ell_1$-norm (L1) and a scaled-$\ell_1$-norm (L1-scaled). The L1 method is a straightforward implementation of logistic regression plus $\ell_1$-norm minimization.
The L1-scaled algorithm combines the logistic function with a scaled-$\ell_1$-norm given by
$R(\beta) := \lambda\onenorm{S\beta}$
for some diagonal scaling matrix $S = \diag(s_1,\dots, s_n)$  
and weighting parameter $\lambda > 0$.  In our tests, we defined $s_i$ as the maximum of $1$ and the minimum cost required to obtain the $i$th feature.  
Although this choice is reasonable, it is also ad-hoc, which is necessarily true for any choice of the scaling matrix $S$. This is a consequence of the fact that it takes a complicated cost structure and represents it by $n$ numbers, which is too simplistic.

Figure~\ref{fig:l1-comparison} compares the tradeoff between financial cost and AUC values of Group, L1, and L1-scaled. (Similar plots could be constructed for test result time and nurse time.)
The reported cost of a model is obtained by post-processing, whereby we sum the costs for the unique set of tests required. Each point in the plot represents a pair $(\$,\text{AUC})$ for some model.  For algorithms L1 and L1-scaled, the 
points were obtained by varying the parameter $\lambda$ over the interval $[10^{-3},10^{-7}]$. For algorithm Group based on the regularizer~\eqref{def:R-relaxed}, we fixed $\lamclock = 10^{-7}$ and let $\lambda_{\$}$ take on the same values as $\lambda$ for algorithms L1 and L1-scaled; this placed different levels of emphasis on \emph{only} the financial cost, which further illustrates the flexibility of our cost-driven structured regularizer.
For all three algorithms we only use tests that have a maximum allowed wait time of $50$ minutes.

First, observe that algorithm L1  performs the worst.  In particular, the cheapest model recovered by algorithm L1 costs \$$129$ and had an AUC of approximately $77.5$. At that same price-point, algorithms L1-scaled and Group were able to obtain AUC values of approximately $84.6$ and $86.1$.  This is not surprising since the $\ell_1$-regularizer used by algorithm L1 causes the most predictive features to be chosen first, \emph{without} any regard to the resulting financial cost. This performance is not surprising and may be used to motivate algorithm L1-scaled.  In essence, L1-scaled  incorporates a \emph{rough} measure of the cost for each feature through the choice of $s_i$, as described above.  Second, Figure~\ref{fig:l1-comparison} shows that our cost-sensitive regularizer significantly outperforms algorithm L1-scaled. Third, observe that a (perhaps) surprisingly high AUC value (approximately $84.5$) may be achieved for models without any financial cost by algorithms L1-scaled and Group. For the prediction of sepsis, this means that although expensive tests produce measurements that  allow for better prediction accuracy, one may still do well without incurring any (additional) financial costs. This observation should be leveraged when implementing screening tools or assessing risk stratification.

\section{Conclusions and discussion}

We designed a structured regularizer that captures
the complex cost structure of many applications.  The 
feature, measurement, test, and caregiver activity hierarchy in
healthcare was used as an example, but we showed how our method can be
used anytime the cost structure can be represented as a
finite-layer boolean circuit.  By building a regularizer that was in
harmony with user's application-specific cost preferences, our 
experiments produced a diverse collection of models.
Moreover, our cost-sensitive regularizer achieved
better prediction accuracy for the same (often lower) cost when
compared to $\ell_1$ or weighted-$\ell_1$ norms commonly used.
We comment that the design of our regularizer must only be
done once up-front for each application, and then may be reused
to answer a host of questions, e.g., through model prediction.

Beyond sepsis, our regularizer applies to many prediction problems in healthcare \cite{bates2014big} including early detection of other potentially preventable conditions, e.g., 
pneumonia, c-diff, and renal failure \cite{Fuller_2009_EstimatingCostsPPC}. 
More broadly, our regularizer is applicable to cost-sensitive prediction problems whose cost-graphs may be represented with a logical \band{} and \bor{} structure associated with boolean circuits. 
In traffic prediction, for example, features (e.g., mean and trend) of the traffic velocity can be computed from streams acquired from sources (e.g., querying crowdsourced GPS devices, pneumatic road tubes, piezo-electric sensors, cameras, and manual counting) at different locations including live event streams~\cite{horvitz2012prediction}.
Considerations for choosing a model include the cost of
acquiring and deploying the sensors, the staff time to maintain the
sensors, and the recurring costs of acquiring traffic, weather and live
event streaming data. Depending on the availability and cost of
resources, one may wish to deploy different models in different
regions.

Although our cost-sensitive regularizer may be used in many important
applications, it has limitations.  Its more accurate modeling of the cost-graph is achieved at the expense of requiring additional computation to construct. Converting a \emph{general} r-layer Boolean circuit to a 3-layer Boolean circuit has complexity 
$O(s^{fr})$, where $s$ is the number of nodes and $f$ is the fan (the largest number of allowed gate inputs/outputs) of the circuit.
However, most cost-graphs are highly structured, thus dramatically
reducing the computational cost.  For example, constructing the regularizer for the ICU application took approximately 10 seconds on a MacBook Air laptop (1.8 GHz Intel Core i5 processor with 4GB of RAM). This modest additional cost is a consequence of the structure of the cost-graph: most nodes have relatively few connections to nodes in adjacent layers, and the logical gates mostly contain simple \bor{} and \band{} constructs.
Since these properties hold for many cost-graphs in practice, our approach is often practical.

\section*{Acknowledgments}

We would like to thank Mu Wei for fruitful conversations that ultimately lead to the ideas presented here.
We would also like to thank Katherine Henry for her help in obtaining and
cleaning the data used in the healthcare example, and for many
conversations in which we benefited from her expert knowledge of the data. Finally, 
we thank Dr. Harold Lehmann for helping us understand end-user preferences in the clinical environment.

%
%
%
%
%
%

\bibliographystyle{named}
\bibliography{references}

\end{document}